\pdfoutput=1

\documentclass[11pt]{article}

\usepackage[final]{acl}
\usepackage{booktabs}
\usepackage{float}

\usepackage{times}
\usepackage{latexsym}
\usepackage{amsmath}
\usepackage{xcolor}

\usepackage[T1]{fontenc}
\usepackage{xcolor}
\usepackage{siunitx}

\usepackage[utf8]{inputenc}

\usepackage{microtype}

\usepackage{inconsolata}
\usepackage{graphicx}
\usepackage[most]{tcolorbox}
\newtcolorbox{boxK}{
    sharpish corners, 
    boxrule = 0pt,
    toprule = 4.5pt, 
    enhanced,
    fuzzy shadow = {0pt}{-2pt}{-0.5pt}{0.5pt}{black!35} 
}
\usepackage{multirow}

\title{Revisiting NLI: Towards Cost-Effective and Human-Aligned Metrics for Evaluating LLMs in Question Answering}

\author{Sai Shridhar Balamurali \\
  University of Illinois at Chicago  \\
  IL, Chicago\\
  \small{
     \href{mailto:sbalam3@uic.edu}{sbalam3@uic.edu}}
   \And Lu Cheng \\
  University of Illinois at Chicago  \\
  IL, Chicago\\
  \small{
     \href{mailto:lucheng@uic.edu}{lucheng@uic.edu}}
 }

\begin{document}
\maketitle
\footnotetext{We acknowledge the use of AI assistants (GPT-4o/Claude 3.7) for editing assistance and improving clarity of expression in portions of this manuscript. All research content, analyses, and scientific contributions were conceived and developed by the authors.}

\begin{abstract}
Evaluating answers from state-of-the-art large language models (LLMs) is challenging: lexical metrics miss semantic nuances, whereas “LLM-as-Judge” scoring is computationally expensive. We re-evaluate a lightweight alternative—off-the-shelf Natural Language Inference (NLI) scoring augmented by a simple lexical-match flag and find that this decades-old technique matches GPT-4o’s accuracy (89.9\%) on long-form QA, while requiring orders-of-magnitude fewer parameters. To test human alignment of these metrics rigorously, we introduce DIVER-QA, a new 3000 sample human annotated benchmark spanning five QA datasets and five candidate LLMs. Our results highlight that inexpensive NLI-based evaluation remains competitive and offer DIVER-QA as an open resource for future metric research. 
\end{abstract}

\section{Introduction}
The rapid advancement of large language models (LLMs) like GPT \cite{brown2020, OpenAI2023}, LLaMA \cite{touvron2023,grattafiori2024llama}, and Gemini \cite{gemini1,gemini1.5} necessitates robust evaluation methodologies. Question Answering (QA) benchmarks like MMLU \cite{hendrycks2020}, MMLU-pro \cite{wang2024mmlu}, and HotpotQA \cite{yang2018hotpotqa} have become standard for assessing model capabilities, but present evaluation challenges. While MMLU relies on multiple-choice questions that can introduce biases \cite{mcq}, HotpotQA expects concise answers that don't accommodate the verbose outputs modern LLMs often generate.

This means that if a model generates the correct answer with different wording it will be considered wrong and could also create a disparity between model classes. Larger models can follow strict instructions to produce concise answers, but smaller models—despite possessing considerable knowledge—often struggle when constrained to restrictive formats. When allowed to generate explanatory responses, these models frequently demonstrate correct reasoning, as shown in this HotpotQA example using Mistral-7B \cite{jiang2023mistral}:

\begin{boxK}
\small
\textbf{Q:} Were Scott Derrickson and Ed Wood of the same nationality? \\
\textbf{Reference:} yes \\
\textbf{Mistral-7B response:} Scott Derrickson is an American Director, while Ed Wood was a American filmmaker. Both are of the same nationality.
\end{boxK}

Current evaluation approaches include: (1) traditional reference-based metrics like Exact Match, which lack semantic understanding; (2) neural model-based metrics \cite{bertscore,bleurt,bem} requiring task-specific training; and (3) computationally expensive "LLM-as-Judge" approaches \cite{judgellm,li2024generation}. None effectively balance accuracy with efficiency when evaluating verbose outputs against reference answers, with each approach introducing potential biases \cite{judgellm,confusion}.

In this paper, we revisit Natural Language Inference (NLI) as a solution to this problem and also experiment augmenting a general NLI model with lexical matching to address substring matching weaknesses. Our weighted logistic regression framework achieves performance comparable to state-of-the-art models while remaining computationally efficient.

Our contributions include:
\begin{itemize}
   \item A diverse QA dataset for testing metrics across modern instruction-tuned LLMs.
   \item Evidence that NLI models can evaluate long-form QA responses nearly as well as GPT-4o (89.9\% accuracy) with significantly lower costs.
   \item Empirical proof of our approach's generalizability and cost-effectiveness across various QA tasks.
\end{itemize}

\section{Background and Related Works}
Traditional lexical metrics for QA, e.g., Exact Match (EM) and F1 score, are inadequate to evaluate contemporary LLMs. These LLMs often generate verbose, instruction-following responses where correct answers are embedded within broader context, leading to unfairly low scores on metrics designed for concise extractions \cite{bertscore, bleurt, bem, gemini1.5, mcq}.

This inadequacy spurred the development of neural metrics like BERTScore \cite{bertscore}, BLEURT \cite{bleurt}, which leveraged contextual embeddings to better capture semantic similarity, and BEM \cite{bem} which used semantic equivalence classification. While an improvement over purely lexical methods, these early neural approaches still exhibited limitations in discerning nuanced semantic equivalence and achieved only modest correlation with human judgment when answers were phrased very differently from references \cite{bleurt}.

More recently, the use of LLMs themselves as evaluators (e.g., "LLM-as-judge" or for direct reference-based assessment) has become state-of-the-art, demonstrating superior semantic understanding and correlation with human ratings \cite{kamalloo2023evaluating, kamalloo2024towards, judgellm}. However, this approach carries a significant drawback: LLM-based evaluators, whether proprietary or open-source, are computationally intensive and prohibitively expensive for many researchers, hindering widespread adoption and reproducibility \cite{confusion}. Furthermore, these LLM judges can introduce their own biases, such as verbosity or positional preferences \cite{judgellm}.

This landscape reveals a critical gap: a need for evaluation metrics that can approach the semantic acuity of powerful LLM evaluators without their substantial computational and financial overhead. Our work addresses this by revisiting and enhancing an efficient neural approach to offer a robust, cost-effective alternative for evaluating modern LLM outputs.

\section{Methodology}
Our research revisits NLI as a foundation for evaluating LLM-generated answers against reference responses. While NLI has been previously explored for QA evaluation \cite{harabagiu2006methods}, we reexamine its effectiveness specifically for verbose outputs from instruction-tuned models and enhance it with complementary signals. 

\subsection{NLI-based Metric}
NLI offers a particularly suitable framework for QA evaluation because it directly addresses the core question: \textit{does the generated answer entail the reference answer? }This formulation naturally handles paraphrasing, additional context, and varied expression—all common in modern LLM outputs.

For our NLI model, we utilize DeBERTa-v3-large-NLI \cite{laurer2024less}, a model specifically built as a universal classifier using NLI. This model was trained on 33 datasets with 389 diverse classes, encompassing 885,242 NLI hypothesis-premise pairs from multiple sources including MultiNLI \cite{williams2018broad}, Fever-NLI \cite{nie2019combining}, Adversarial-NLI (ANLI) \cite{nie2020adversarial}), LingNLI \cite{parrish2021does}, and WANLI \cite{liu2022wanli}. The authors demonstrated that NLI can serve as a universal classification task following similar principles as instruction fine-tuning of generative LLMs, making it particularly well-suited for zero-shot classification tasks like QA evaluation.

We selected this model specifically for its generalized zero-shot performance, which enables researchers to use it "out of the box" without requiring additional fine-tuning for specific QA datasets. This choice aligns with our goal of providing computationally efficient evaluation methods that remain accessible to researchers with limited resources. Given a question $q$, a candidate answer $a$ and a reference answer $r$, we format inputs as 
\begin{boxK}
    "[CLS] question: $q$ answer: $a$ [SEP] question: $q$ ground truth: $r$ [SEP]" 
\end{boxK}
\noindent to provide the model with full context, taking the entailment probability as our semantic score. Entailment probability provides the most informative measure of semantic equivalence, as contradiction merely indicates opposing propositions without confirming similarity, while neutrality signifies an absence of inferential relationship. Within the tripartite probability distribution, entailment uniquely quantifies the degree to which one statement logically implies another.

\begin{figure}
    \centering
    \includegraphics[width=1.0\linewidth]{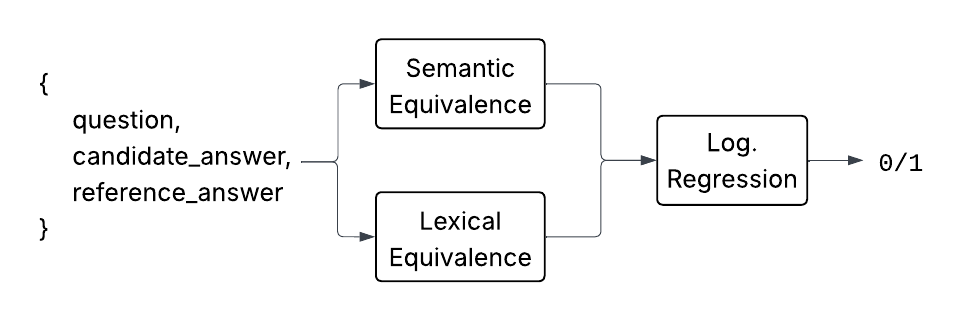}
    \caption{Framework for the NLI+lex model.}
    \label{fig:framework}
\end{figure}

\subsection{NLI Augmented with Lexical Equivalence}
While NLI excels at capturing semantic equivalence (\texttt{se(q,a,r)}), it can occasionally falter with precise factual assessments where an answer is explicitly stated but embedded within verbose context. To address this, we augment NLI with a complementary lexical matching signal (\texttt{lm(a,r)}), a simple binary function indicating if the reference \texttt{r} appears as a substring in the answer \texttt{a}. These two signals are linearly combined:
\begin{equation*}
z = w_1 \cdot \texttt{se(q, a, r)} + w_2 \cdot \texttt{lm(a, r)}.
\end{equation*}
The weights ($w_1$, $w_2$) are learned through logistic regression, yielding a probability of correctness, $P(\texttt{correct}) = 1 / (1 + e^{-z})$. Responses are then classified as correct (1) or incorrect (0) using a 0.5 threshold (Figure \ref{fig:framework}. We can calibrate these weights using any QA dataset. More details on our implementation is present in Appendix \ref{sec:model_training}). We refer to this hybrid approach as \textbf{NLI+lex} which strategically leverages NLI's semantic strengths while efficiently addressing its limitations in factual recall, offering a generalizable evaluation method that avoids the data leakage concerns inherent in training QA-specific evaluation models.

\section{Evaluation}
\subsection{Setup}
\label{sec:Evaluation_setup}
We compare our NLI-based approach against three categories of evaluation metrics: (1) \textbf{traditional lexical metrics}, including ROUGE-L and F-1 score; (2) \textbf{neural semantic metrics}, such as BERTscore, BLEURT, and BEM; and (3) \textbf{LLM-as-judge} approaches, specifically GPT-4o \cite{hurst2024gpt} and Llama-3.1-8B-Instruct \cite{grattafiori2024llama}. We will refer to GPT-4o, acting as an evaluator, as \textit{GPT-Eval} and Llama-3.1-8B-Instruct as \textit{Llama-Eval}. As a collective, we will refer to the above models as evaluators for clarity. To see details on these evaluators, refer to Appendix \ref{sec:metric_info}.

We measure each metric's alignment with human judgment using \textbf{Accuracy, F-1 score, }and \textbf{Matthews Correlation Coefficient (MCC)} which we will refer to as \textbf{our metrics} when talked about collectively during our evaluation. We employ MCC as our primary correlation metric, as it is optimally suited for assessing performance on binary classification tasks, which constitutes the fundamental predictive function performed by the models in this study. MCC is described as below: \\
\begin{equation}
\small
\begin{aligned}
      &\text{MCC} = \\
    &\frac{TP \cdot TN - FP \cdot FN}{\sqrt{(TP + FP)(TP + FN)(TN + FP)(TN + FN)}}, 
\end{aligned}
\end{equation}
\\
where \(TP\): true positives, \(TN\): true negatives, \(FP\): false positives, \(FN\): false negatives. 

Additionally, we analyze computational requirements to demonstrate our method's efficiency advantage over LLM-based approaches while maintaining comparable performance. we use \textbf{active parameter count} as a primary proxy for the amount of compute. While not perfectly proportional, such orders-of-magnitude differences in parameters for models of similar dense architectures strongly indicate a correspondingly large reduction in computational cost, far exceeding what typical hardware or software optimizations could bridge. In the case of GPT-4o, we assume it uses around 200B active params based on \cite{epoch2024frontier}.

\subsection{Dataset}
While datasets with human evaluations like those from \cite{wang2024evaluating,bem} are available, their limited diversity of LLMs and QA types restricts the ability to robustly evaluate the general performance of different metrics. Also, models in \cite{bem} do not produce answers with the same verbosity as present day LLMs. To thoroughly assess metric effectiveness, we curated DIVER-QA (DIverse Question Answering) evaluation dataset, sampling 120 questions from each of five varied sources: AdversarialQA \cite{bartolo2020beat} which offers challenging questions designed to be difficult for models, SQuAD \cite{rajpurkar2018know} which provides extractive QA from Wikipedia articles, MedQA \cite{jin2021disease} which includes specialized medical domain questions, HotpotQA \cite{yang2018hotpotqa} which requires multi-hop reasoning and synthesis, and TriviaQA \cite{joshi2017triviaqa} which features fact-based questions from quiz competitions. 

We generated answers using five models of varying sizes and architectures: claude-3.5-sonnet \cite{claude35sonnet}, llama-3-70B, llama-3-8B \cite{grattafiori2024llama}, and phi-3-mini \cite{abdin2024phi}, yielding 3,000 question-answer pairs. Human annotators assessed each response, determining whether it matched the reference answer with a simple "yes" or "no" judgment. Final human evaluations were determined by majority vote. More details on dataset construction can be found in Appendix \ref{sec:data collection}. In our evaluation, we shall refer to the models that generated the answers in DIVER-QA as \textit{candidate models }for clarity.

\subsection{Results and Discussion}

\begin{table}[htbp]
  \centering
    \begin{tabular}{l
                  S[table-format=1.2]
                  S[table-format=1.6]
                  S[table-format=1.2]}
    \toprule
    \textbf{Evaluator} & \textbf{Accuracy} & \textbf{F1-score} & \textbf{MCC} \\
    \midrule
    F1-score              &  0.2853    &   0.0377 & 0.0727 \\
    ROUGE-L               &  0.2860       &   0.0420 & 0.0619 \\
    BERTscore             &  0.4876    &   0.4834 & 0.2453 \\
    BLEURT                &  0.4877    &   0.4834 & 0.2453 \\
    Llama-Eval &  0.5623    &   0.6784 & 0.0042 \\
    BEM                   &  0.8230       &   0.8781 & 0.5549 \\
    NLI                   &  0.8357    &   0.8778 & 0.6536 \\
    GPT-Eval                 &  0.8360       & 0.8797 & 0.6408 \\
    \textbf{NLI + lex}         & \textbf{0.8450}  & \textbf{0.8865} & \textbf{0.6603} \\
    \bottomrule
  \end{tabular}
  \caption{Performance of Evaluators compared to Human Evaluation on DIVER-QA. The MCC provides the correlation and Accuracy and F1 are the agreement with human evaluation. }
  \label{tab:performance}
\end{table}

From Table \ref{tab:performance}, we can see that traditional metrics such as F1-score and ROUGE-L vastly underperform when it comes to evaluating long-form answer generation. BERTscore, BLEURT and Llama-Eval scores are midrange; however, Llama-Eval also had the worst MCC. BEM, NLI GPT-4o (GPT-Eval) and our model tend to perform well with accuracy above 80\%. Among these four, BEM tends to fall behind in terms of MCC by almost 10\%. NLI+lex edges out GPT-4o by around 1\% across all metrics. While both models perform strongly, NLI+lex consistently outperforms the base NLI model by 1-3\% across metrics. The lexical matching component addresses NLI's occasional weakness in capturing exact factual matches within verbose responses, providing complementary information that enhances evaluation robustness. Deeper Analysis including reasoning for LLama-Eval's poor performance is present in Appendix \ref{sec:further_results}. \\

\begin{figure}[h]
    \centering
    \includegraphics[width=1.0\linewidth]{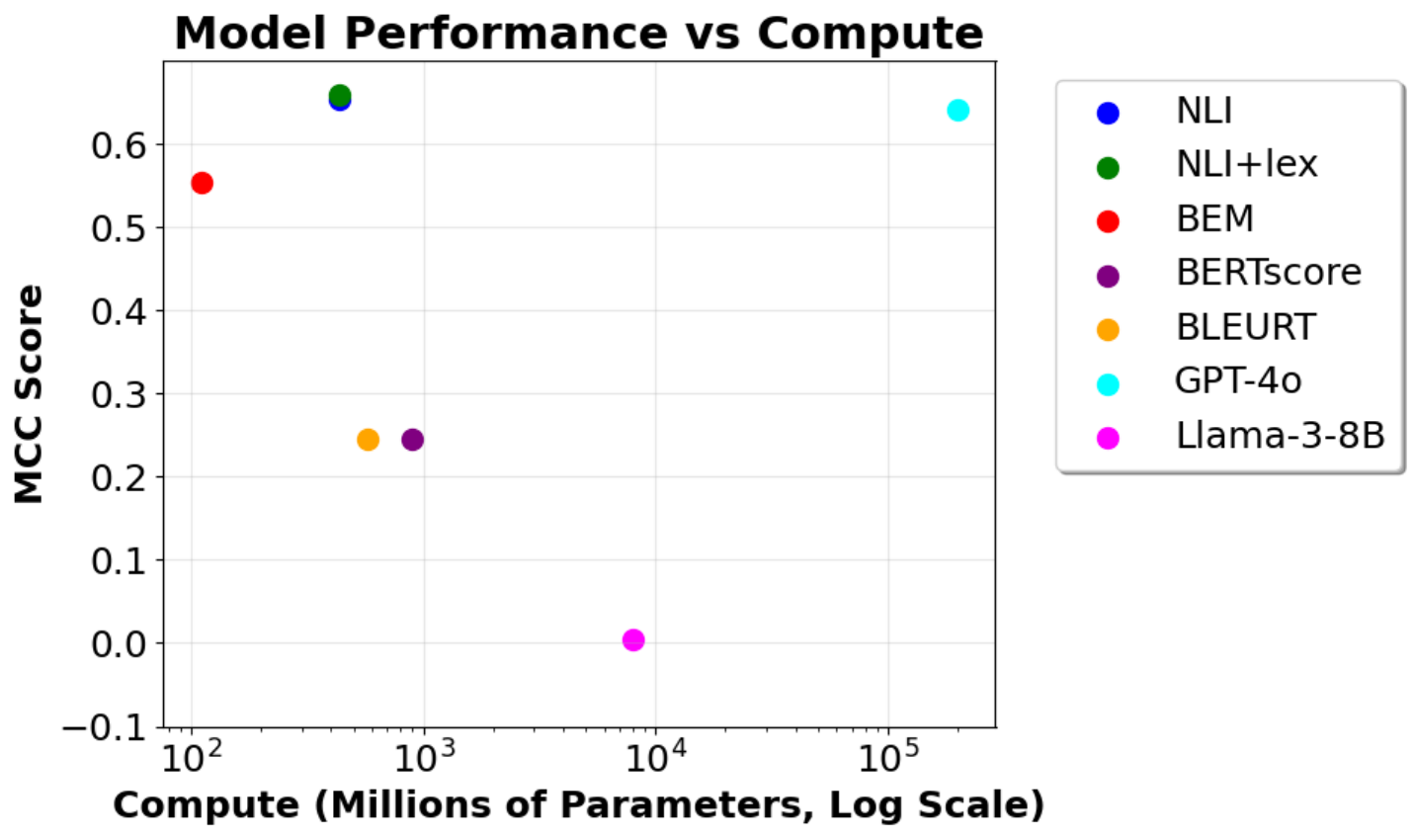}
    \caption{Compute vs Performance ratio of the metrics used.}
    \label{fig:per-com}
\end{figure}

Next, we examine the relationship between computational cost and performance, as depicted in Fig.\ref{fig:per-com}, with the aim of identifying a model that provides high performance while minimizing computational expense. Such a model would allow users to efficiently evaluate their datasets. From Fig.\ref{fig:per-com}, we observe that both the NLI model and our proposed NLI+lex model achieve performance comparable to GPT-4o, yet they incur the second-lowest computational costs. Although the BEM model has the lowest computational cost, it displays a 10\% reduction in MCC as discussed earlier, compared to GPT-4o. Hence, the NLI and NLI+lex models offer the optimal trade-off between computational cost and predictive performance.

\section{Conclusion}
We show that an off-the-shelf NLI model, boosted with a single lexical-match flag, can judge long-form QA answers as accurately as GPT-4o (89.9\% on our DIVER-QA benchmark) while using several orders of magnitude fewer parameters. This provides an alternative to heavyweight “LLM-as-Judge'’ systems.
To spur further work, we release DIVER-QA, a human-annotated dataset covering five QA tasks and five modern LLMs. Together, our method and benchmark offer a lightweight, transparent test bed for developing new metrics.
Although NLI+lex is strong on factual QA, it still misses subtleties such as partially correct or multi-step answers. Future efforts should extend the approach to dialogue, multimodal inputs, and finer-grained judgments. 
\section*{Limitations}
While our study provides valuable insights into metric performance across several domains, the current evaluation datasets represent a focused subset of possible question types. Expanding dataset diversity across additional domains would further strengthen the generalizability of our findings. Exploring alternative prompting strategies for the LLM-as-a-judge based approaches could enhance their performance, however they could also affect the compute cost. Further analysis here will help us get a better grasp on the tradeoffs of different evaluation methods. Exploring alternate methods of augmenting the NLI model or using more sophisticated lexical equivalence component could help strengthen our case. Lastly, incorporating mathematical and algorithmic reasoning tasks represents a natural extension that would complement our existing methodology and broaden its applicability to additional important domains.

\bibliography{custom}

\appendix

\section{Details on Data Collection}
\label{sec:data collection}
 To construct the DIVER-QA dataset, we used the distribution of questions shown in Table \ref{tab:dataset_distribution}. While the selection process was not randomized, it was guided by the goal of capturing a diverse range of question types within each dataset. For example, in the case of SQuAD, which includes questions from a variety of scientific and academic domains (e.g., physics, chemistry, mathematics, and computer science), we ensured that our sample included representation from these different areas. We applied a similar approach to the other datasets, selecting the first instances that met our desired distribution of question types. Given the structured nature and topical breadth of these datasets, this method yielded a diverse and representative evaluation set.

 \begin{table}
\centering
\begin{tabular}{lc}
\hline
\textbf{Dataset} & \textbf{No. of questions} \\
\hline
AdversarialQA & 120 \\
SQuAD & 120 \\
MedQA & 120 \\
HotpotQA & 120 \\
TriviaQA & 120 \\
\hline
\textbf{Total} & 600 \\
\hline
\end{tabular}
\caption{Distribution of questions across datasets in DIVER-QA}
\label{tab:dataset_distribution}
\end{table}
 
\subsection{Human Annotation Methodology
}
To establish a gold standard for evaluating model-generated answers, we employed a human annotation process involving five evaluators out of which 3 were Computer Science Graduate students and two were high-school students. Given the difficulty of the task, there was no need for any expertise. Our primary annotation goal was to determine if a model-generated answer ``matches the ground truth'' for a given question. 
A prediction matches the ground truth if:
\begin{enumerate}
    \item The message conveyed is equivalent despite using different words
    \item In case of a name, place, or numeric value (e.g., number of people or items), it correctly identifies the entity from the context
\end{enumerate}

\begin{figure*}[t]
    \centering
    \includegraphics[width=1.0\linewidth]{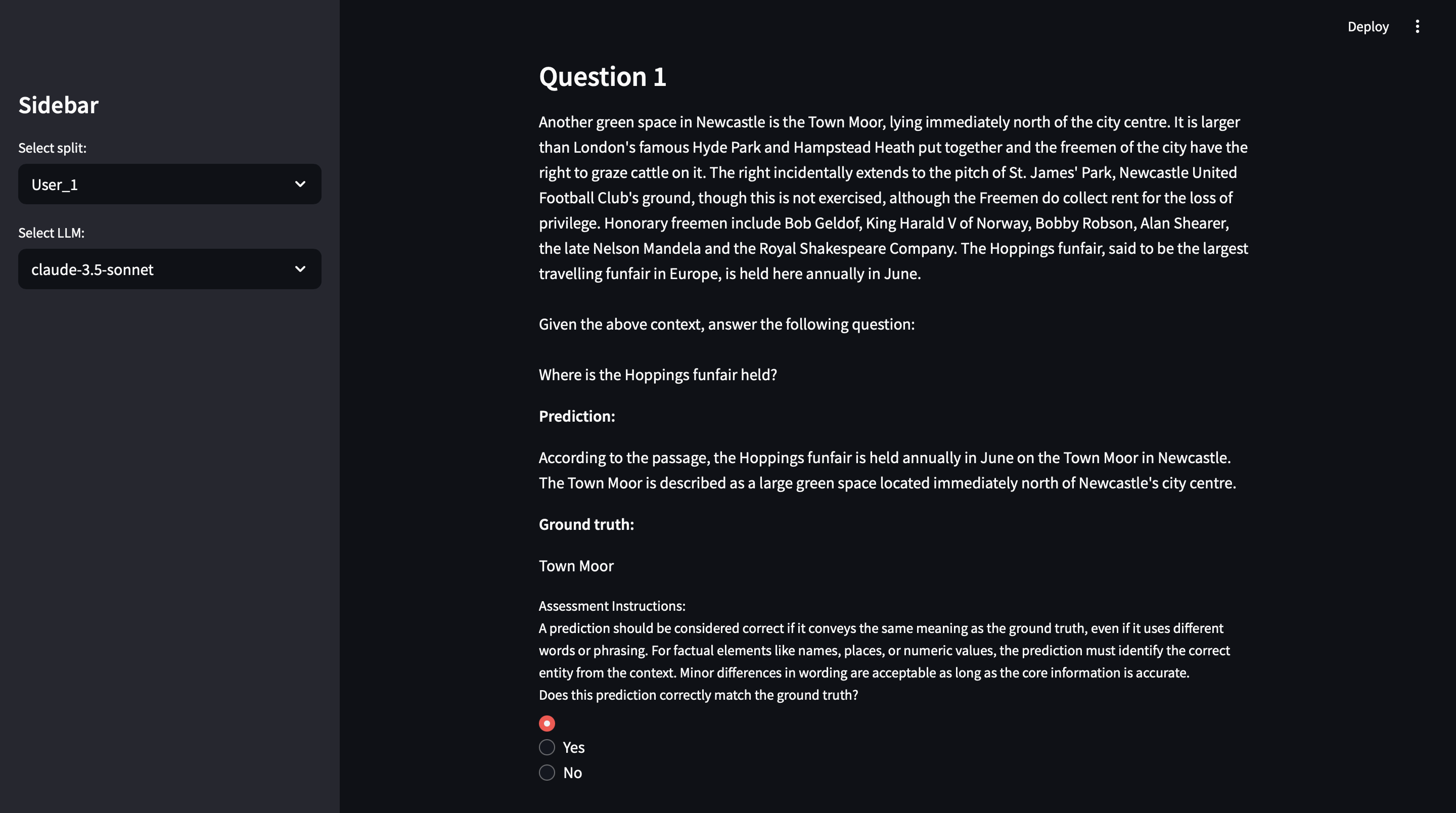}
    \caption{Human Annotator UI}
    \label{fig:UI}
\end{figure*}

The UI used by the evaluators is show in \ref{fig:UI}
\subsection{Annotation Distribution}
To ensure each of the 3,000 questions received judgments from three distinct evaluators, we implemented a structured workload distribution scheme. The dataset was divided into it's five subdatasets ($d_1$ to $d_5$). Let's refer to each of these subdatasets as partitions. Each of the five evaluators was assigned a unique combination of three partitions, resulting in each evaluator assessing 1,800 question-answer pairs. The specific assignments were as follows:

\begin{table}[h]
\centering
\begin{tabular}{ll}
\hline
\textbf{Evaluator} & \textbf{Assigned Partitions} \\
\hline
Evaluator 1 & $d_1$, $d_2$, $d_4$ \\
Evaluator 2 & $d_1$, $d_3$, $d_5$ \\
Evaluator 3 & $d_1$, $d_3$, $d_5$ \\
Evaluator 4 & $d_2$, $d_3$, $d_4$ \\
Evaluator 5 & $d_2$, $d_4$, $d_5$ \\
\hline
\end{tabular}
\caption{Distribution of evaluation partitions among evaluators}
\label{tab:evaluator_assignments}
\end{table}

The final human evaluation label for each question-answer pair was determined by a majority vote among the three evaluators.

\subsection{Inter-Annotator Agreement}
To assess the reliability of our human judgments, we calculated inter-annotator agreement (IAA) on the subsets of data where evaluator assignments overlapped. We report Matthews Correlation Coefficient (MCC) and simple percentage agreement (Accuracy) for all evaluator pairs. Tables \ref{tab:iaa_set1}, \ref{tab:iaa_set2}, \ref{tab:iaa_set3}, \ref{tab:iaa_set4} and \ref{tab:iaa_set5} show the IAA scores for the five partitions.

Across all models and evaluator pairs, the simple percentage agreement was consistently high, generally exceeding 95\%, indicating a strong baseline level of consistency. Where rating variance allowed for robust calculation, MCC and Cohen's Kappa values also demonstrated substantial chance-corrected agreement.

In some instances, particularly for certain models (e.g., Claude-3.5-Sonnet outputs in Set 2), MCC values were 0. This is a known characteristic of these metrics when observed agreement is very high but concentrated on a single rating category (e.g., most answers rated as 'correct'), leading to minimal variance in the ratings for that subset. In such cases, the corresponding simple accuracy remained exceptionally high, confirming evaluator consensus rather than a lack of agreement.

\begin{table}[h]
\centering
\small
\begin{tabular}{llcc}
\hline
\textbf{Model} & \textbf{Evaluator Pair} & \textbf{MCC} & \textbf{Accuracy} \\
\hline
\multirow{3}{*}{Claude-3.5-Sonnet} 
 & (1,2) & 1.000 & 1.000 \\
 & (1,3) & 0.922 & 0.983 \\
 & (2,3) & 0.922 & 0.983 \\
\hline
\multirow{3}{*}{Mixtral-8x7B} 
 & (1,2) & 0.908 & 0.975 \\
 & (1,3) & 0.965 & 0.992 \\
 & (2,3) & 0.877 & 0.967 \\
\hline
\multirow{3}{*}{Llama-3-8b} 
 & (1,2) & 0.973 & 0.992 \\
 & (1,3) & 0.834 & 0.950 \\
 & (2,3) & 0.857 & 0.958 \\
\hline
\multirow{3}{*}{Llama-3-70b} 
 & (1,2) & 0.934 & 0.983 \\
 & (1,3) & 0.889 & 0.975 \\
 & (2,3) & 0.900 & 0.975 \\
\hline
\multirow{3}{*}{Phi-3-mini} 
 & (1,2) & 0.937 & 0.983 \\
 & (1,3) & 0.937 & 0.983 \\
 & (2,3) & 1.000 & 1.000 \\
\hline
\end{tabular}
\caption{Inter-Annotator Agreement for Partition 1}
\label{tab:iaa_set1}
\end{table}

\begin{table}[h]
\centering
\small
\begin{tabular}{llcc}
\hline
\textbf{Model} & \textbf{Evaluator Pair} & \textbf{MCC} & \textbf{Accuracy} \\
\hline
\multirow{3}{*}{Claude-3.5-Sonnet} 
 & (1,2) & 0.000 & 0.992 \\
 & (1,3) & 0.000 & 0.992 \\
 & (2,3) & 0.000 & 1.000 \\
\hline
\multirow{3}{*}{Mixtral-8x7B} 
 & (1,2) & 0.813 & 0.992 \\
 & (1,3) & 0.624 & 0.975 \\
 & (2,3) & 0.768 & 0.983 \\
\hline
\multirow{3}{*}{Llama-3-8b} 
 & (1,2) & 0.000 & 0.992 \\
 & (1,3) & 1.000 & 1.000 \\
 & (2,3) & 0.000 & 0.992 \\
\hline
\multirow{3}{*}{Llama-3-70b} 
 & (1,2) & 0.000 & 0.983 \\
 & (1,3) & 0.704 & 0.992 \\
 & (2,3) & 0.000 & 0.992 \\
\hline
\multirow{3}{*}{Phi-3-mini} 
 & (1,2) & 0.862 & 0.992 \\
 & (1,3) & 0.698 & 0.975 \\
 & (2,3) & 0.809 & 0.983 \\
\hline
\end{tabular}
\caption{Inter-Annotator Agreement for Partition 2}
\label{tab:iaa_set2}
\end{table}

\begin{table}[h]
\centering
\small
\begin{tabular}{llcc}
\hline
\textbf{Model} & \textbf{Evaluator Pair} & \textbf{MCC} & \textbf{Accuracy} \\
\hline
\multirow{3}{*}{Claude-3.5-Sonnet} 
 & (1,2) & 0.982 & 0.992 \\
 & (1,3) & 0.982 & 0.992 \\
 & (2,3) & 1.000 & 1.000 \\
\hline
\multirow{3}{*}{Mixtral-8x7B} 
 & (1,2) & 0.983 & 0.992 \\
 & (1,3) & 0.983 & 0.992 \\
 & (2,3) & 1.000 & 1.000 \\
\hline
\multirow{3}{*}{Llama-3-8b} 
 & (1,2) & 1.000 & 1.000 \\
 & (1,3) & 1.000 & 1.000 \\
 & (2,3) & 1.000 & 1.000 \\
\hline
\multirow{3}{*}{Llama-3-70b} 
 & (1,2) & 1.000 & 1.000 \\
 & (1,3) & 1.000 & 1.000 \\
 & (2,3) & 1.000 & 1.000 \\
\hline
\multirow{3}{*}{Phi-3-mini} 
 & (1,2) & 1.000 & 1.000 \\
 & (1,3) & 1.000 & 1.000 \\
 & (2,3) & 1.000 & 1.000 \\
\hline
\end{tabular}
\caption{Inter-Annotator Agreement for Partition 3}
\label{tab:iaa_set3}
\end{table}

\begin{table}[h]
\centering
\small
\begin{tabular}{llcc}
\hline
\textbf{Model} & \textbf{Evaluator Pair} & \textbf{MCC} & \textbf{Accuracy} \\
\hline
\multirow{3}{*}{Claude-3.5-Sonnet} 
 & (1,2) & 0.951 & 0.975 \\
 & (1,3) & 0.935 & 0.967 \\
 & (2,3) & 0.983 & 0.992 \\
\hline
\multirow{3}{*}{Mixtral-8x7B} 
 & (1,2) & 0.934 & 0.967 \\
 & (1,3) & 0.932 & 0.967 \\
 & (2,3) & 0.966 & 0.983 \\
\hline
\multirow{3}{*}{Llama-3-8b} 
 & (1,2) & 0.933 & 0.967 \\
 & (1,3) & 0.765 & 0.883 \\
 & (2,3) & 0.795 & 0.900 \\
\hline
\multirow{3}{*}{Llama-3-70b} 
 & (1,2) & 0.917 & 0.958 \\
 & (1,3) & 0.917 & 0.958 \\
 & (2,3) & 0.901 & 0.950 \\
\hline
\multirow{3}{*}{Phi-3-mini} 
 & (1,2) & 0.950 & 0.975 \\
 & (1,3) & 0.934 & 0.967 \\
 & (2,3) & 0.949 & 0.975 \\
\hline
\end{tabular}
\caption{Inter-Annotator Agreement for Parition 4}
\label{tab:iaa_set4}
\end{table}

\begin{table}[h]
\centering
\small
\begin{tabular}{llcc}
\hline
\textbf{Model} & \textbf{Evaluator Pair} & \textbf{MCC} & \textbf{Accuracy} \\
\hline
\multirow{3}{*}{Claude-3.5-Sonnet} 
 & (1,2) & 0.908 & 0.975 \\
 & (1,3) & 0.877 & 0.967 \\
 & (2,3) & 0.895 & 0.975 \\
\hline
\multirow{3}{*}{Mixtral-8x7B} 
 & (1,2) & 0.915 & 0.975 \\
 & (1,3) & 0.711 & 0.908 \\
 & (2,3) & 0.733 & 0.917 \\
\hline
\multirow{3}{*}{Llama-3-8b} 
 & (1,2) & 0.839 & 0.942 \\
 & (1,3) & 0.794 & 0.917 \\
 & (2,3) & 0.738 & 0.892 \\
\hline
\multirow{3}{*}{Llama-3-70b} 
 & (1,2) & 0.821 & 0.950 \\
 & (1,3) & 0.822 & 0.942 \\
 & (2,3) & 0.719 & 0.908 \\
\hline
\multirow{3}{*}{Phi-3-mini} 
 & (1,2) & 0.850 & 0.942 \\
 & (1,3) & 0.819 & 0.917 \\
 & (2,3) & 0.837 & 0.925 \\
\hline
\end{tabular}
\caption{Inter-Annotator Agreement for Partition 5}
\label{tab:iaa_set5}
\end{table}

The robust IAA scores provide confidence in the human-generated labels used as the gold standard in our subsequent analyses.

\section{Model Training Details}
\label{sec:model_training}
Since the logistic regression model was trained exclusively on the outputs derived from the Natural Language Inference (NLI) model and our lexical equivalence function, the specific dataset used to generate questions and answers is unlikely to exert a substantial downstream influence. This limited effect occurs because the logistic regressor does not directly utilize information from the questions and answers themselves; instead, it evaluates only the degree of semantic and lexical similarity between reference and candidate responses to determine correctness. Consequently, we employed a separate subset consisting of 5,000 questions from the HotpotQA dataset for training the logistic regressor. 

\section{Evaluator Selection Details}
\label{sec:metric_info}
We provide detailed information regarding the specific models and sources utilized for each evaluator in our comparative analysis in Table \ref{tab:metrics}. For BERTScore, we employed the deberta-v2-xl-mnli model, which was identified as the best-performing model according to its official GitHub model listings. For BLEURT, we selected BLEURT-20, as this variant exhibited superior performance among available BLEURT models. The F1-score was computed using the standard formula. To facilitate comparison between the continuous outputs of evaluation metrics and our binary human annotations, we applied a standardized binarization process. A threshold of 0.5 was established; values exceeding this threshold were classified as 1 (correct), while those below were classified as 0 (incorrect). This normalization approach enabled direct comparative analysis between continuous metric scores and discrete human judgments. For GPT-4o and Llama-3.1-8B-Instruct, we accessed the respective APIs as indicated in Table \ref{tab:metrics}. Both models employed constrained decoding \cite{geng2023grammar}, also described as structured outputs in their API documentation. Specifically, the models were constrained to produce a binary output—'1' if the candidate matched the reference and '0' otherwise.

\begin{table}
\centering
\begin{tabular}{|l|l|}
\hline
\textbf{Evaluator} & \textbf{Source} \\
\hline
BERTscore & Official Github \\
\hline
BLEURT & Official Github\\
\hline
BEM & HF/kortukov \\
\hline
NLI & HF/MoritzLaurer \\
\hline
ROUGE-L & Pytorch Ignite\\
\hline
F1-score & Self-implemented \\
\hline
GPT-4o & OpenAI API \\
\hline
Llama-3.1-8B-Instruct & Fireworks API \\
\hline
\end{tabular}
\caption{Metrics and their sources. HF refers to hugging face implementations and the names after the forward slash represent the HF author name.}
\label{tab:metrics}
\end{table}

\section{Further Results}
\label{sec:further_results}
\subsection{Model-wise Analysis}
For the model-wise analysis, we consider the performance of an evaluator on evaluating each candidate model's answers across all the sub-datasets in DIVER-QA. The idea is that this will allow us to understand any variations in performance of the metrics due to the model being evaluated. 
\begin{figure}[h]
    \centering
    \includegraphics[width=1.0\linewidth]{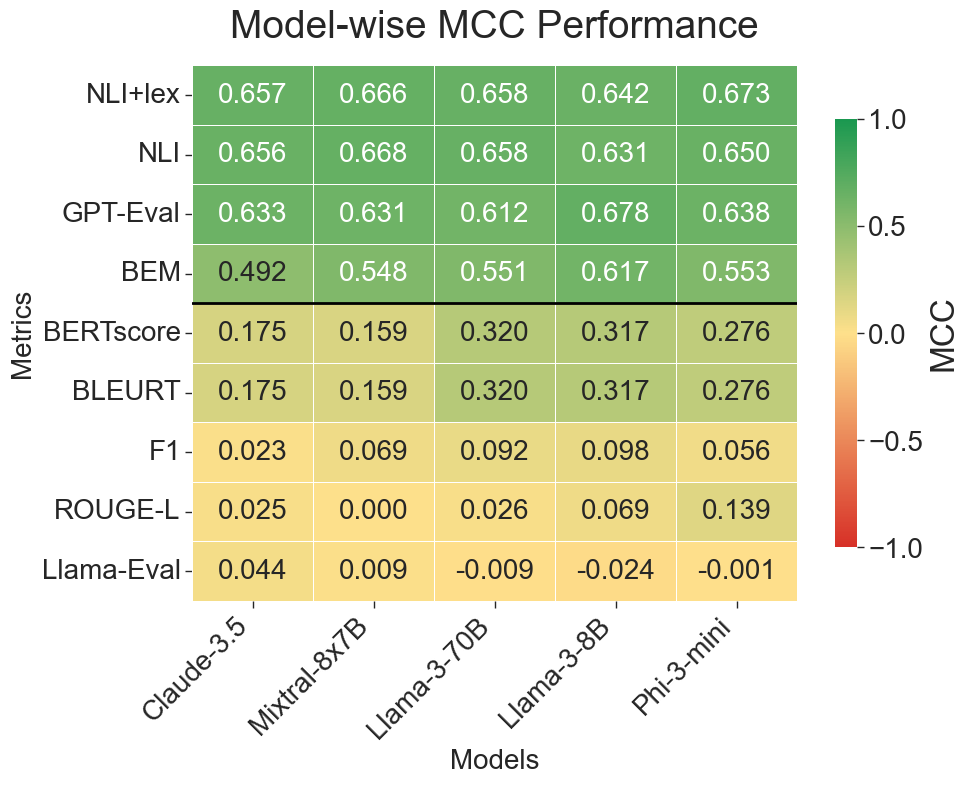}
    \caption{Modelwise MCC scores}
    \label{fig:model_mcc}
\end{figure}

\begin{figure}[h]
    \centering
    \includegraphics[width=1.0\linewidth]{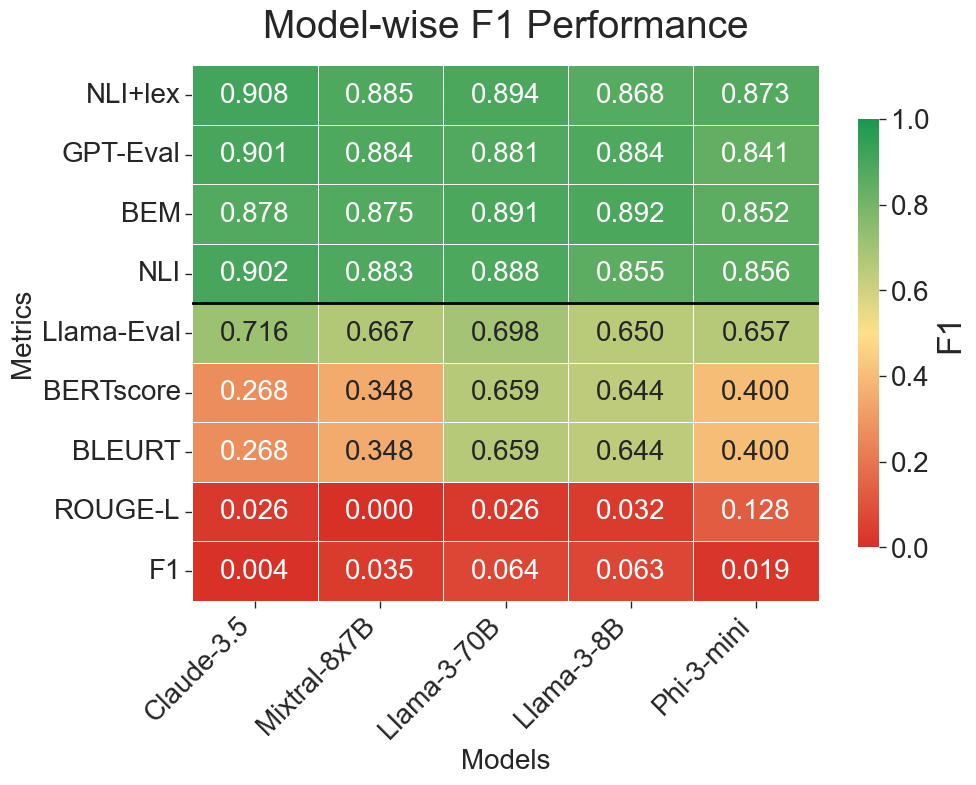}
    \caption{Modelwise F1 scores}
    \label{fig:model_f1}
\end{figure}

\begin{figure}[h]
    \centering
    \includegraphics[width=1.0\linewidth]{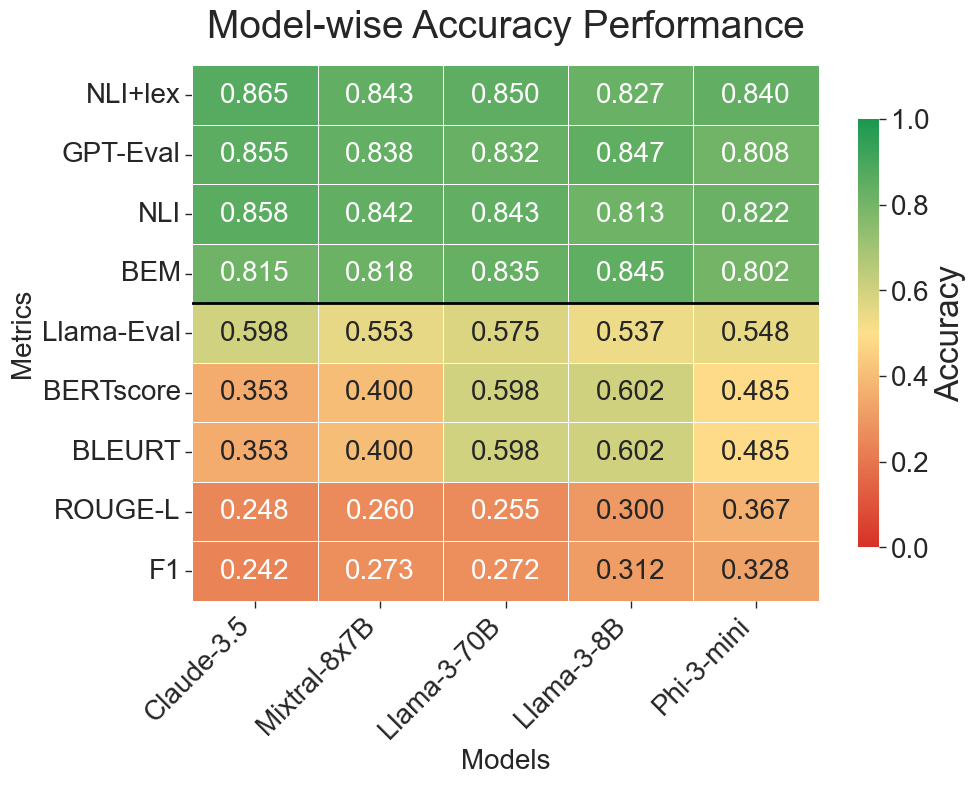}
    \caption{Model-wise Accuracy scores}
    \label{fig:model_acc}
\end{figure}

The model-wise analysis reveals distinct performance hierarchies and notable variances among evaluation metrics across the five evaluated models, as illustrated in Figures \ref{fig:model_mcc},\ref{fig:model_f1},\ref{fig:model_acc}.
NLI+lex consistently demonstrates superior performance across all metrics and evaluated models. For MCC (Figure \ref{fig:model_mcc}), it achieves correlations between 0.642-0.673, maintaining its lead in F1 scores (Figure \ref{fig:model_f1}) with values from 0.868-0.908, and in accuracy measurements (Figure \ref{fig:model_acc}) ranging from 0.827-0.865. This consistent performance establishes NLI+lex as the most reliable metric for alignment with human evaluation.\\
The base NLI model follows as a strong performer, particularly in MCC correlation where it nearly matches NLI+lex, with values between 0.631-0.668. However, an interesting pattern emerges in F1 and accuracy measurements, where NLI occasionally underperforms compared to other metrics. For instance, when evaluating Llama-3-8B, GPT-4o achieves higher accuracy (0.847 vs. 0.813) and competitive F1 scores.\\
A particularly notable observation is Llama-Eval's performance disparity across metrics: it exhibits moderate accuracy but near-zero or negative MCC values across all models. Our investigation revealed this stems from architectural limitations when constraining the model to single-token binary outputs (0/1) without intermediate generation steps. While statistical biases allow Llama-Eval to achieve moderate accuracy, the near-zero or negative MCC values indicate its predictions lack meaningful correlation with ground truth when accounting for the complete confusion matrix structure. \\
This finding highlights smaller parameter models' fundamental dependency on generative reasoning pathways that cannot be circumvented through constrained decoding.\\
Interestingly, GPT-Eval doesn't suffer from these same constraints, performing exceptionally well despite constrained decoding. It achieves the highest MCC correlation (0.678) when evaluating Llama-3-8B and maintains strong performance across most models. This resilience stands in stark contrast to Llama-Eval's performance, suggesting that larger, more sophisticated models can effectively evaluate responses even with constrained outputs.\\
BEM occupies the middle tier, showing moderate but consistent performance across all metrics. While it achieves respectable MCC correlations (0.492-0.617) and F1 scores (0.852-0.892), it experiences a notable 10\% drop in MCC performance compared to top-performing metrics, resulting in scores around 55\% correlation with human judgment. This significant performance gap highlights the superior evaluation capabilities of NLI-based approaches, particularly our NLI+lex model, which consistently demonstrates better alignment with human evaluation across all tested models.\\
The remaining metrics (BERTscore, BLEURT, ROUGE-L, and F1) consistently underperform across all evaluation dimensions, suggesting limited utility for sophisticated evaluation tasks. Their poor correlation with human judgment indicates that syntax-based or traditional metrics lack the semantic understanding necessary for nuanced evaluation.

\subsection{Dataset-wise Analysis}
For the dataset-wise analysis, we consider the performance of an evaluator on evaluating all candidate model answers for each sub-dataset in DIVER-QA. This will allow us to understand any variations in performance of the metrics due to the dataset being used for evaluation of the candidate models. 
\begin{figure}[h]
    \centering
    \includegraphics[width=1.0\linewidth]{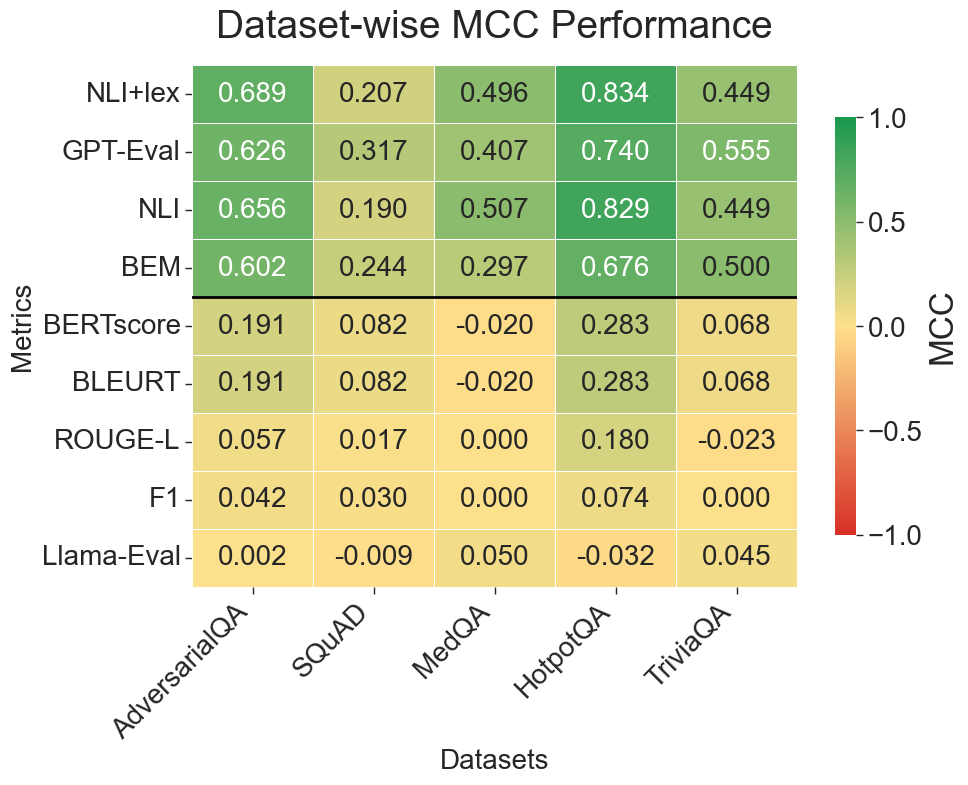}
    \caption{Datasetwise MCC scores}
    \label{fig:dataset_mcc}
\end{figure}

\begin{figure}[h]
    \centering
    \includegraphics[width=1.0\linewidth]{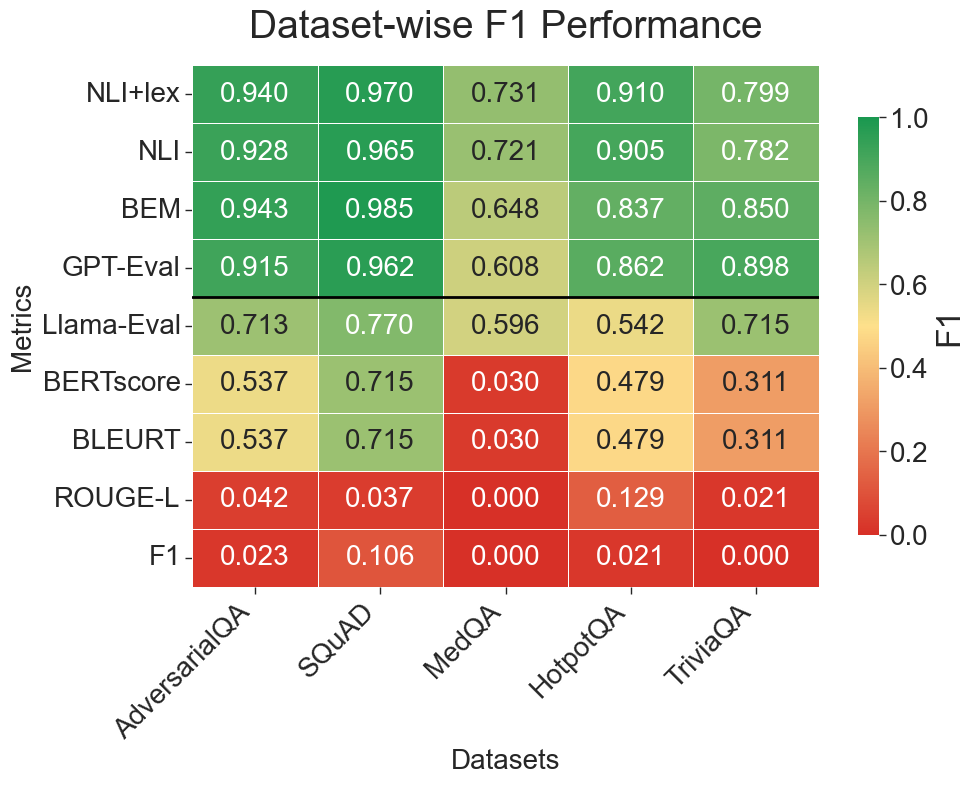}
    \caption{Datasetwise F1 scores}
    \label{fig:dataset_f1}
\end{figure}
\begin{figure}[h]
    \centering
    \includegraphics[width=1.0\linewidth]{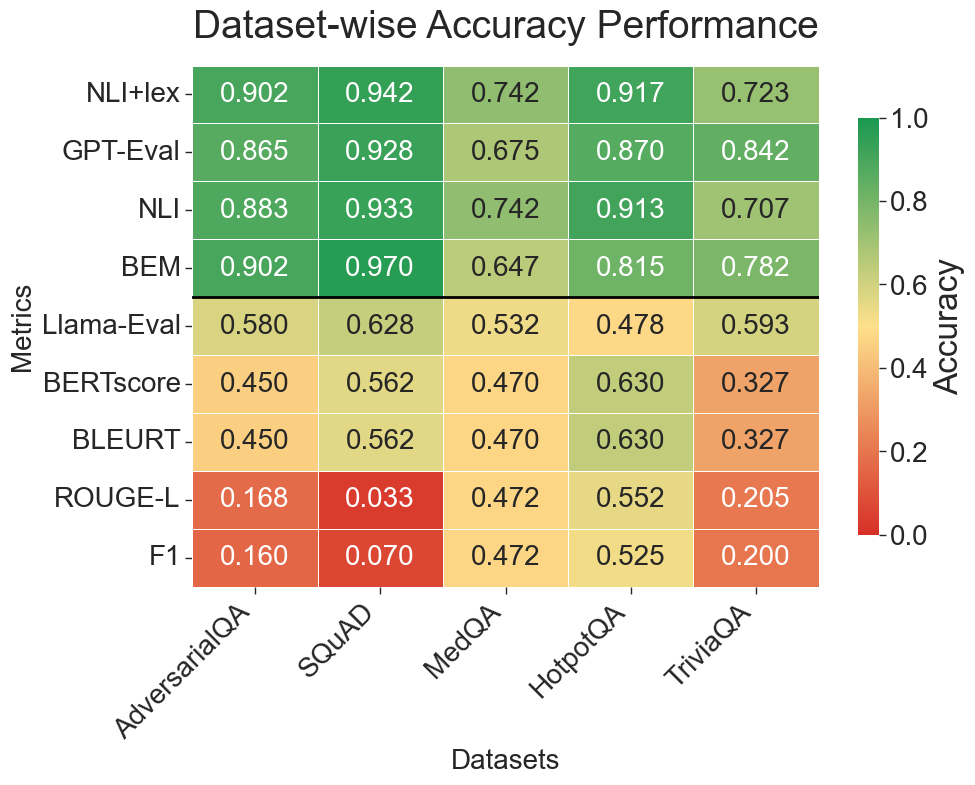}
    \caption{Datasetwise Accuracy scores}
    \label{fig:dataset_acc}
\end{figure}

 The dataset-wise analysis, as illustrated in Figures \ref{fig:dataset_mcc}, \ref{fig:dataset_f1} and \ref{fig:dataset_acc},
 reveals significant performance variations across multiple evaluation metrics and datasets, providing insights into how different metrics handle diverse question-answering domains.\\
Domain specificity creates dramatic performance differences, particularly evident in MedQA, where F1 scores range from 0.000 to 0.731 across metrics. BERTscore and BLEURT demonstrate significant limitations, achieving merely 0.030 F1 scores on MedQA. A critical factor in the performance gap between these metrics and the top 4, across different datasets, is the lack of access to the question context, evaluating answers solely through comparison between reference and candidate answers.\\
SQuAD presents an interesting case study in metric training influence. BEM achieves its highest F1 score (0.985) on SQuAD, directly attributable to its training on SQuAD data. This highlights how evaluation metrics can be biased toward datasets similar to their training data.\\
An intriguing statistical pattern emerges across datasets: the apparent contradiction between high accuracy but relatively weak MCC scores on SQuAD and MedQA. This occurs because MCC becomes less informative when predictions show little variation. For these datasets, Accuracy and F1 provide more reliable performance indicators, confirming strong performance on SQuAD (accuracy > 0.9 for top-performing metrics) while showing more moderate performance on MedQA (0.65-0.74 range).\\
The identical values between BERTscore and BLEURT across all datasets in the Accuracy and F1 charts are not a methodological error but rather an artifact of our evaluation protocol. Both metrics produce continuous scores in the [0,1] range that we binarize for classification purposes, resulting in identical binary outcomes despite their underlying continuous scores being different.\\
Regarding Llama-Eval's poor MCC performance, we observe the same pattern across datasets as discussed in the modelwise analysis, with the model showing moderate accuracy but near-zero or negative MCC values due to architectural limitations with constrained decoding.\\
The zero F1 scores (0.000) for ROUGE-L and F1 metric on MedQA are not methodological errors but result from score clipping at four decimal places. \\
A clear performance hierarchy emerges across datasets: metrics with access to the question (NLI+lex, NLI, GPT-Eval, and BEM) consistently outperform those without question access (BERTscore, BLEURT, ROUGE-L, and F1). This pattern holds particularly for complex reasoning tasks in AdversarialQA and HotpotQA, where NLI-based approaches demonstrate superior performance, suggesting that natural language inference with question context is particularly effective for evaluating complex reasoning and multi-hop question answering.
\end{document}